\documentclass[12pt]{elsarticle}

\usepackage{amssymb}
\usepackage{amsmath}
\journal{ArXiv}

\usepackage{natbib}

\begin{document}
\begin{frontmatter}

%% Title, authors and addresses

%% use the tnoteref command within \title for footnotes;
%% use the tnotetext command for theassociated footnote;
%% use the fnref command within \author or \address for footnotes;
%% use the fntext command for theassociated footnote;
%% use the corref command within \author for corresponding author footnotes;
%% use the cortext command for theassociated footnote;
%% use the ead command for the email address,
%% and the form \ead[url] for the home page:
%% \title{Title\tnoteref{label1}}
%% \tnotetext[label1]{}
%% \author{Name\corref{cor1}\fnref{label2}}
%% \ead{email address}
%% \ead[url]{home page}
%% \fntext[label2]{}
%% \cortext[cor1]{}
%% \affiliation{organization={},
%%             addressline={},
%%             city={},
%%             postcode={},
%%             state={},
%%             country={}}
%% \fntext[label3]{}

\title{(LiFT) Lightweight Fitness Transformer: A language-vision model for Remote Monitoring of Physical Training}

%% use optional labels to link authors explicitly to addresses:
%% \author[label1,label2]{}
%% \affiliation[label1]{organization={},
%%             addressline={},
%%             city={},
%%             postcode={},
%%             state={},
%%             country={}}
%%
%% \affiliation[label2]{organization={},
%%             addressline={},
%%             city={},
%%             postcode={},
%%             state={},
%%             country={}}

\author[inst1]{Alexander Postlmayr}

\affiliation[inst1]{organization={Dept. of Electrical and Computer Engineering University of California, San Diego}%Department and Organization
            % addressline={Address One}, 
            % city={City One},
            % postcode={00000}, 
            % state={State One},
            % country={Country One}
            }
\author[inst1]{Pamela Cosman}
\author[inst1]{Sujit Dey}

% \affiliation[inst2]{organization={Department Two},%Department and Organization
%             addressline={Address Two}, 
%             city={City Two},
%             postcode={22222}, 
%             state={State Two},
%             country={Country Two}}

\begin{abstract}
We introduce a fitness tracking system that enables remote monitoring for exercises using only a RGB smartphone camera, making fitness tracking more private, scalable, and cost effective. Although prior work explored automated exercise supervision, existing models are either too limited in exercise variety or too complex for real-world deployment. Prior approaches typically focus on a small set of exercises and fail to generalize across diverse movements. In contrast, we develop a robust, multitask motion analysis model capable of performing exercise detection and repetition counting across hundreds of exercises, a scale far beyond previous methods. We overcome previous data limitations by assembling a large-scale fitness dataset, {\it Olympia}, covering more than 1,900 exercises. To our knowledge, our vision-language model is the first that can perform multiple tasks on skeletal fitness data. On {\it Olympia}, our model can detect exercises with $76.5\%$ accuracy and count repetitions with $85.3\%$ off-by-one accuracy, using only RGB video. By presenting a single vision-language transformer model for both exercise identification and rep counting, we take a significant step toward democratizing AI-powered fitness tracking.
\end{abstract}

% %%Graphical abstract
% \begin{graphicalabstract}
% \includegraphics[scale=0.5]{VPT-Architecture.png}
% \end{graphicalabstract}

% %%Research highlights
% \begin{highlights}

% \item Enabling motion-text learning using software-only
% \item Using a single machine learning model to supervise human exercises with action recognition and repetitive action counting
% \item Mobile exergaming with camera-only
% \end{highlights}

% \begin{keyword}
% %% keywords here, in the form: keyword \sep keyword
% Computer vision \sep Machine learning \sep Mobile computing \sep Musculoskeletal system \sep Patient monitoring \sep Patient rehabilitation \sep Smart healthcare 
% \end{keyword}

\end{frontmatter}

%% \linenumbers

%% main text
\section{Introduction}
\label{sec:introduction}

Although musculoskeletal disorders are among the main drivers of healthcare costs in the United States \cite{dieleman_us_2020}, and physical exercise \cite{zadro_overcoming_2020,zadro_physical_2019} is a well-established intervention for managing these conditions, proper technique is crucial for effective rehabilitation and injury prevention. However, access to professional fitness training remains costly and typically requires in-person supervision, limiting its accessibility. Research shows that strength training can reduce inflammation \cite{schwalm_activation_2015} and even reduce the risk of certain cancers \cite{mctiernan_physical_2019}. In addition, muscle mass is inversely correlated with mortality risk \cite{abramowitz_muscle_2018}, highlighting the importance of maintaining physical fitness. As modern lifestyles become increasingly sedentary, it is critical to reintegrate structured exercise into daily life. Our research aims to accelerate this shift by democratizing access to personalized fitness training through AI-driven scalable solutions.

% -------- this section needs to talk about how important 3D HPE
Physical training with AI-powered remote monitoring would allow people to exercise alone and yet have some of the benefits that come from the supervision of a professional fitness trainer. Using a computer, exercises are shown to a user,  so that they can exercise anytime, anywhere. 
 Virtual physical training needs to provide personalized feedback \cite{lambert_app_2017}, as current online programs are inadequate, especially for tracking user exercises, repetitions, and correct posture \cite{mitchell_quality_2022}. Accountability through tracking improves motivation \cite{uzawa_outcome_2018, hammer_mechanical_2007}, and fitness programs with goal setting, reminders, and progress tracking are more successful \cite{friedrich_combined_1998}. 
 
Current technological approaches to exercise feedback show diverse methods.
Many existing works \cite{milanko_liftright_2020,liao_deep_2020,yu_skeleton-based_2021, elkholy_efficient_2020, qu_llms_nodate} rely on expensive specialized hardware, such as RGB-D cameras, limiting their use. 
% \cite{yu_skeleton-based_2020} uses UI-PRMD RGB-D data, \cite{elkholy_efficient_2020} KINECT \cite{qu_llms_nodate} NTU+D dataset
% % \cite{milanko_liftright_2020} - IMU
Software-only solutions offer a scalable alternative, using monocular videos from standard RGB cameras found in smartphones and laptops. Here, motion analysis is performed either directly on the video pixels or on skeletal key points extracted from the video using 3D Human Pose Estimation (3D HPE) models.  Video-only methods have shown sufficient results \cite{levy_live_2015,chung_long_2022,zhang_repetitive_2021,zhang_context-aware_2020, tang_multicounter_2024,hu_transrac_2022}, however, they are inefficient for mobile devices. Streaming video to the cloud drastically reduces privacy, consumes significant bandwidth, and introduces network latency. Addressing these drawbacks, recent work in 3D HPE models \cite{bazarevsky_blazepose_2020,zhu_motionbert_2023} has demonstrated the ability to extract human poses from video in real time on mobile devices. Crucially, this approach provides physics-constrained human joint angles, a primary metric in motion analysis. Irrelevant background information is discarded. Finally, using 3D HPE enables privacy-preserving deep learning approaches at scale by allowing the efficient storage of motion information.

% Some works have relied on 2D HPE 
% OpenPose:
% \cite{ferreira_transformers_2022}, \cite{sinclair_puioio_2023}, 
% \cite{hsu_viewpoint-invariant_2021}, 

\begin{figure}
    \includegraphics[width=\linewidth]{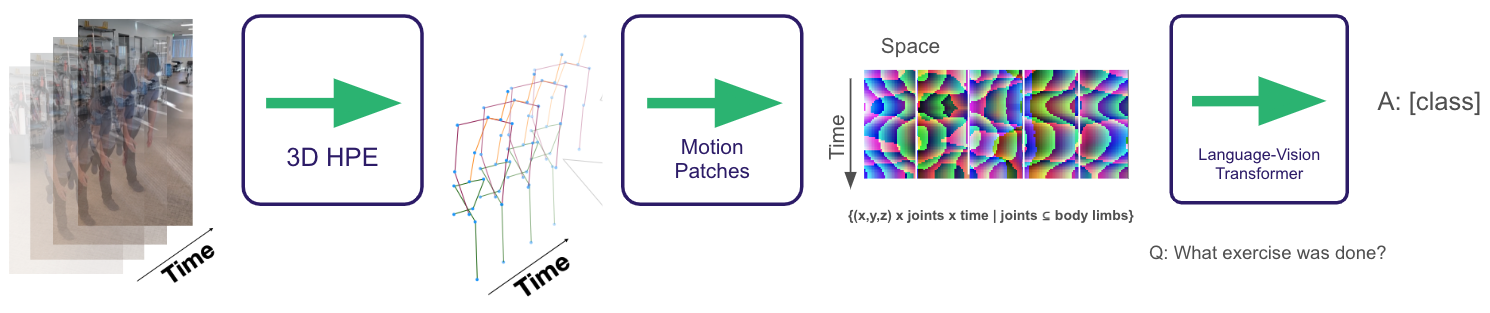}
    \caption{The overview of LiFT where RGB videos are converted to skeletal keypoints using 3D HPE. The skeletal motion is converted into motion patches, which are then given to the language-vision model along with a question task. The output of the model is a series of possible classes which belong to the motion.  }\label{fig:overview} 
 %   \vspace{-5mm}
  \end{figure}

In this paper, we present our deep learning approach for multitask motion analysis. LiFT is a {\bf Li}ghtweight {\bf F}itness {\bf T}ransformer capable of exercise detection and repetition counting on many exercises. The model overview is shown in Figure ~\ref{fig:overview}. Our approach is mobile-friendly and can be used for at-home personal fitness tracking. Unlike existing solutions, LiFT is trained on multiple exercise datasets that contain many exercises seen in real-world applications. Table \ref{lift_datasets} shows existing fitness datasets along with the number of subjects and exercises included. Our training dataset, {\it Olympia}, is described in Section \ref{olympia}.  Our contributions are as follows:
\begin{itemize}
\item{LiFT- the first 3D HPE skeleton based language-vision model capable of performing exercise detection and repetition counting across hundreds of exercises.} %, a scale previously not attempted.}

\item{We demonstrate that our approach effectively captures fine-grained relationships between motion segments and textual descriptions, unlike methods that rely on global feature alignment. Through evaluations on four publicly available datasets and few-shot fine-tuning experiments, we show that our model generalizes to previously unseen exercise classes, addressing a key challenge in real-world fitness applications.}

\item{We assemble the {\it Olympia} dataset, which contains 7,618 annotated videos for exercise detection and repetition counting.}
\end{itemize}

\section{Related Work}
As will be discussed in detail in the following, existing fitness analysis models often attempt to handle multiple tasks, but are typically limited to fewer than 30 exercises and rely on handcrafted features or template-based methods, limiting their ability to generalize to diverse movements. Additionally, most motion-language models are built on complex architectures trained on broad, general motion datasets, making them difficult to adapt for specialized applications like exercise tracking. In contrast, our motion-language approach is designed specifically for exercises, offering a more efficient, scalable, and real-world-ready solution without excessive model complexity.

\subsection{3D HPE Skeletal-based Methods} 
We define the tasks required for evaluation of fitness exercises as: 1) exercise detection (classification), 2) repetitive action counting, and 3) exercise repetition correctness assessment.

% \begin{table}
% \caption[Literature Fitness Datasets]{Comparison of fitness exercise datasets found in the literature.}
% \label{table:datasets} 
% \renewcommand\tabularxcolumn[1]{>{\RaggedRight\arraybackslash}p{#1}}
% \begin{tabularx}{.9\linewidth}{lccc}
% \toprule
% \multicolumn{1}{X}{Dataset Name}
% &\multicolumn{1}{X}{Exercises}
% &\multicolumn{1}{X}{Subjects }
% &\multicolumn{1}{X}{Modality}
% \midrule
% WorkoutSU-10 & $10$ & $12$ & $RGB-D$\\
% PHYTMO & $6$ & $30$ & $MoCap$\\
% NOL-18 & $18$ & $6$ & $RGB$\\
% KIMORE & $5$ & $72$ & $RGB, RGB-D$\\
% UI-PRMD & $10$ & $10$ & $RGB, RGB-D, MoCap$\\
% Fit3D  & $47$ & $8$ & $RGB, RGB-D, MoCap$\\
% Repcount  & $6$ & $\leq 1041$ & $RGB$\\
% \bottomrule
% Olympia \footnotemark~ & $\mathbf{1,749}$ & $\mathbf{>71}$ & $RGB$\\

% \end{tabularx}
% \end{table}

\begin{table}[h]
\centering
\caption[Physical Fitness Datasets for HAR and RAC]{Comparing existing datasets for exercise detection and repetition counting.}
\label{lift_datasets}
\begin{tabular}{|p{100pt}|p{50pt}|}
% \begin{tabular}{|m{0.25\linewidth} | m{0.14\linewidth} | m{0.14\linewidth} |}
\hline
\textbf{Dataset Name} & \textbf{Exercises} \\
\hline
% WorkoutSU-10 & $10$ & $12$ & $RGB-D$\\
PHYTMO \cite{garcia-de-villa_database_2022} & $6$ \\
NOL-18 \cite{cheng_periodic_2023} & $18$\\
KIMORE \cite{capecci_kimore_2019} & $5$ \\
UI-PRMD \cite{vakanski_data_2018} & $10$ \\
Fit3D \cite{fieraru_aifit_2021}  & $47$ \\
Repcount \cite{hu_transrac_2022} & $6$\\
\hline\hline
Olympia  & $\mathbf{1306}$ \\
\hline
\end{tabular}
\end{table}

% \begin{table}[h]
% \centering
% \caption[Physical Fitness Datasets for HAR and RAC]{Comparing existing datasets for exercise detection and repetition counting.}
% \label{lift_datasets}
% \begin{tabular}{|p{100pt}|p{50pt}|p{50pt}|p{150pt}|}
% % \begin{tabular}{|m{0.25\linewidth} | m{0.14\linewidth} | m{0.14\linewidth} |}
% \hline
% \textbf{Dataset Name} & \textbf{Exercises} & \textbf{Subjects} & \textbf{Modality}\\
% \hline
% % WorkoutSU-10 & $10$ & $12$ & $RGB-D$\\
% PHYTMO \cite{garcia-de-villa_database_2022} & $6$ & $30$ & $MoCap$\\
% NOL-18 \cite{cheng_periodic_2023} & $18$ & $6$ & $RGB$\\
% KIMORE \cite{capecci_kimore_2019} & $5$ & $72$ & $RGB, RGB-D$\\
% UI-PRMD \cite{vakanski_data_2018} & $10$ & $10$ & $RGB, RGB-D, MoCap$\\
% Fit3D \cite{fieraru_aifit_2021}  & $47$ & $8$ & $RGB, RGB-D, MoCap$\\
% Repcount \cite{hu_transrac_2022}  & $6$ & $\leq 1041$ \footnotemark~ & $RGB$\\
% \hline\hline
% Olympia \footnotemark~ & $\mathbf{1,749}$ & $\mathbf{>71}$ & $RGB$\\
% \hline
% \end{tabular}
% \end{table}

%
% \footnotetext[1]{The number of subjects is not disclosed. Videos were collected by searching for exercises using online sources. Videos may or may not feature the same subject more than once. Therefore the number of subjects is less than or equal to the number of videos.}

% \footnotetext[2]{71 distinct subjects are represented in our Olympia dataset; however, the dataset also contains 100s of synthetic videos where the subjects cannot be individually counted.}

{\bf Single-task:} For software-only methods, several 3D HPE based models exist which focus on only one of these tasks: exercise detection \cite{noauthor_unlabeled_nodate,gajbhiye_ai_2022}, repetition counting \cite{hu_transrac_2022, dwibedi_counting_2020, postlmayr_personalpt_2024}, or quality assessment \cite{garg_short_2024}.

{\bf Multi-task:} 
Several existing models attempt to perform multiple fitness analysis tasks, but they are significantly limited in scope, typically evaluated on fewer than 30 exercises and relying on handcrafted features or template-based methods that do not generalize well to diverse movements. The authors in \cite{francisco_computer_2023} use a multi-layer perceptron trained to classify three exercises as being performed correctly or incorrectly. This approach relies on precomputed Euclidean joint angles, making it difficult to scale to a broader range of movements. In \cite{jaiswal_using_2023}, a random forest classifier classifies four exercises; their Fourier transform-based repetition counting and per-exercise template for form assessment require manual template creation, which does not scale well beyond the predefined exercises. Similarly, in \cite{nair_online_2024}, predefined exercise templates are used for seven exercises to perform recognition and repetition counting. 
%meaning that new exercises require additional template definitions, limiting adaptability. 
The work in \cite{fieraru_aifit_2021} assesses repetition quality using self-similarity metrics, comparing user motion to a trainer template. However, their approach does not perform exercise recognition and is restricted to 30 exercises. The authors in \cite{cheng_periodic_2023} evaluate exercise recognition and repetition counting on 18 exercises by dividing each movement into unit actions (contraction, relaxation, pause, noise) using joint angle thresholds. Their method requires manually defined heuristics for each exercise. In contrast to these approaches, our model leverages motion-language learning to generalize across hundreds of exercises without requiring handcrafted templates or per-exercise tuning.

\subsection{Transformer Based Methods}

 % \cite{yu_exploring_2024} Vision Transformers (mention again in methods)

% \cite{chung_long_2022} video based method (already mentioned in intro)
 Supervised large language models (LLMs) have been shown to be able to learn relationships between natural language and human motion \cite{jiang_motiongpt_2023, chen_motionllm_2024, zhou_avatargpt_2024}. However, the LLMs used (Flan-T5-Base \cite{raffel_exploring_2023}, Vicuna/LLaMA 7B-v1.5  \cite{noauthor_vicuna_nodate,touvron_llama_2023}) require significant compute resources.  They use a token-based approach, where a motion skeleton sequence is encoded into a sequence of motion tokens using a vector quantized variational autoencoder \cite{oord_neural_2018}. The associated natural language label is converted to text tokens \cite{mikolov_efficient_2013,sennrich_neural_2016,song_fast_2021}. A LLM is then used for sequence-to-sequence modeling, to learn the relationships between motion tokens and text tokens. The model is trained for causal inference, enabling it to perform both motion-to-text and text-to-motion generation. This means that, given a sequence of motion inputs, the model can generate a corresponding textual description of the exercise being performed (motion-to-text) and vice versa. These works \cite{jiang_motiongpt_2023, chen_motionllm_2024, zhou_avatargpt_2024} use HumanML3D \cite{guo_ericguo5513humanml3d_2025}, a skeleton-based dataset consisting of general human movement (walking, sitting, etc.) recorded using a motion capture system. Each motion has been annotated with a caption describing the movement. However, this dataset does not include a variety of fitness exercises. Furthermore, the skeletons in HumanML3D are extensively pre-processed. The angular velocities for all skeletal joints are extracted using inverse kinematics for each video, and then applied to a single predetermined skeleton using forward kinematics to create a dataset of uniform skeletons performing different motions. Finally, the skeletons are geometrically aligned to face x,y,z=(0,0,1) at t=0. Overall, existing motion-language methods rely on complex architectures trained on finely curated, general motion datasets. In contrast, our approach is more efficient, tailored for fitness exercises, and optimized for real-world applicability without excessive model complexity.

\section{Methods}

\subsection{Overview}

As shown in Figure \ref{fig:overview}, we first extract 3D keypoints from RGB video using MotionBERT \cite{zhu_motionbert_2023} and combine all the skeletons obtained from each video into a single ``image''. The model learns the relationship between the skeletal motion and natural language labels by minimizing the multiclass cross-entropy loss function of Eq.~(\ref{eq:ce_loss}). During training, each batch consists of both tasks of repetition counting and exercise detection. The trainable model consists of (a) a linear projection layer, (b) the vision transformer, and (c) the classification head.

\subsection{Motion Images}
\label{motion_patches}

\begin{figure}
    \includegraphics[width=\linewidth]{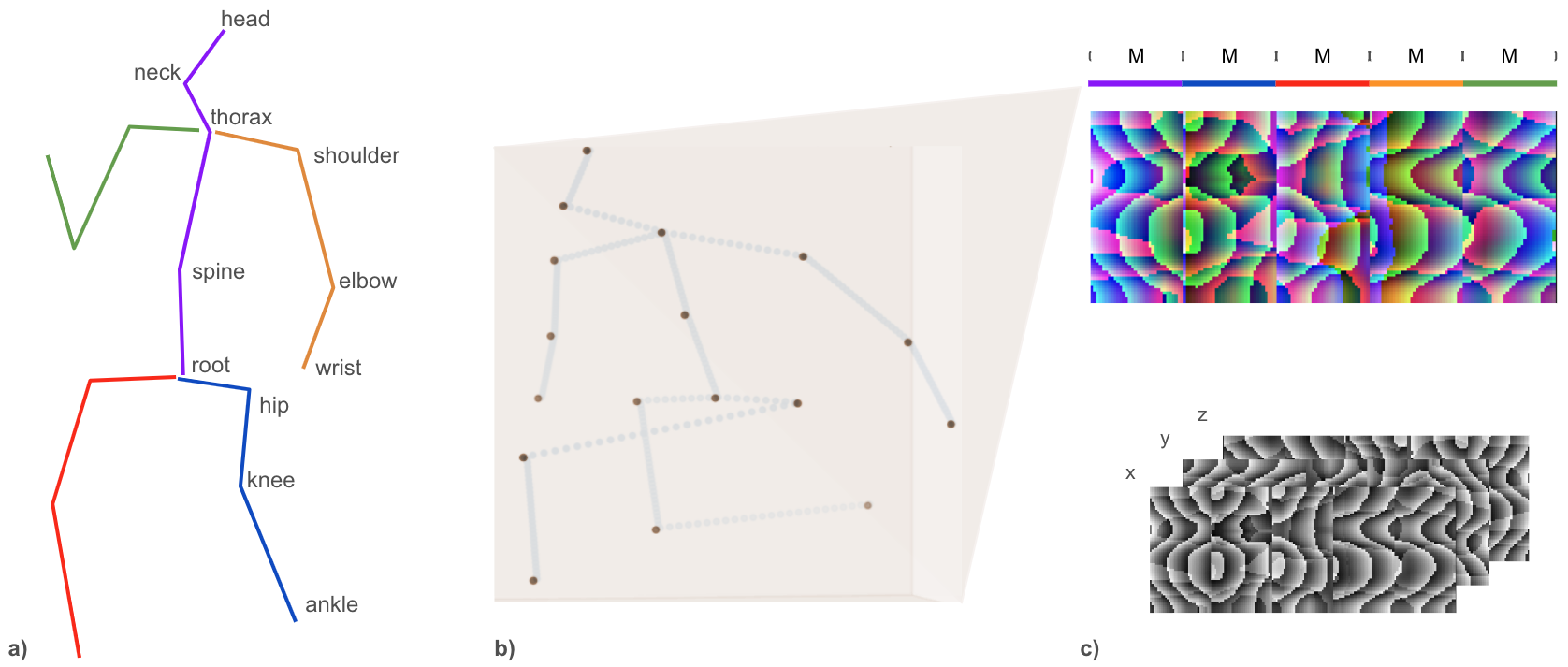}
    \caption{(a) Keypoints in Human3.6M format as detected by MotionBERT; (b) Resulting skeleton after interpolating each kinematic chain to have M=64 positions; (c) Resulting image after concatenating all kinematic chains together (top) while the x,y,z positions are preserved (bottom).  }\label{fig:motion_patches} 
 %   \vspace{-5mm}
  \end{figure}

Similar to \cite{yu_exploring_2024}, we convert skeletal motion into a motion image. Let \( \mathbf{S} \) be the skeleton data, a tensor of shape \( (T, N, 3) \), where \( T \) is the number of time steps, \( N \) is the number of joints, and each joint has a 3D position \( (x, y, z) \). 
Our work uses Human3.6M \cite{ionescu_human36m_2014} format, which has \( N=17 \)  keypoints, as shown in Fig.~\ref{fig:motion_patches}a. To reduce high-frequency noise not observed in human motion,
% ($\geq 5 Hz$),
we average across time with a window size \( w  = 3\), so
%we achieve a half-power, cutoff frequency of 3.21 Hz. 
for each joint \( n \) and dimension \( d \in \{x, y, z\} \), the data is temporally averaged with the values before and after. 
%The first and last frame in the motion are averaged with their adjacent frame only. 
This results in smoothed skeleton data \( \mathbf{S'} \).
% \[
% S'_{t, n, d} = \frac{1}{w} \sum_{i=-\frac{w-1}{2}}^{\frac{w-1}{2}} S_{t+i, n, d}
% \]
% \( \mathcal{C} = \{ C_1, C_2, C_3, C_4, C_5 \} \), where where each chain \( C_i \) 
This is divided into 5 kinematic chains, each involving a subset of the 17 joints. We define the kinematic chains as: [root, spine, thorax, neck, head], [root, left hip, left knee, left ankle], [root, right hip, right knee, right ankle], [thorax, left shoulder, left elbow, left wrist], [thorax, right shoulder, right elbow, right wrist]. A kinematic chain is defined as \( \mathbf{C}(t) = [ \mathbf{n}_1, \mathbf{n}_2, \dots, \mathbf{n}_P ] \), to denote the joint positions at time $t$, where \( P \) is the number of joints in the chain and each \( \mathbf{n}_i \in \mathbb{R}^3 \). For each kinematic chain, we perform 1D linear interpolation on each spatial dimension (x,y,z) to create $\mathbf{C}'(t) = [ \mathbf{n}_1, \mathbf{n}_2, \dots, \mathbf{n}_M ]$, with $M=64$ evenly spaced points, as shown in Fig.~\ref{fig:motion_patches}b. % We interpolate each chain to calculate \( \mathbf{C}'(t) = [ \mathbf{n}'_1, \mathbf{n}'_2, \dots, \mathbf{n}'_{M} ] \) where \( M = 64 \) is the desired number of interpolated points. We first compute the cumulative Euclidean distances for each chain \(r_j = \sum_{j=2}^{P} \|\mathbf{n}_j - \mathbf{n}_{j-1}\|\), for $j=2$ to $P$ and set \(r_1 = 0\). We normalize the cumulative distances \( \left[ r_1, r_2, \dots, r_P \right]\) by dividing by the final value \(r_P \), resulting in normalized positions \( \tilde{r} =  \left[ r_1, r_2, \dots, r_P \right] * \frac{1}{r_p}\). We then define 64 evenly spaced points in the interval \( [0,1] \), denoted by \( \tilde{\mathbf{r}'} \). 
% Let $\bf{C_d}$ be the trajectory over the coordinate 
% $\bf{d}$ over the original points $\tilde{r}$. We define the piecewise linear interpolation function \( \hat{f}_d : [0,1] \to \mathbb{R} \) for each spatial coordinate, \( \hat{f}_d = \text{interp1d}\left( \tilde{r}, C_d \right) \).
% % \( \hat{f}_d(\tilde{r'},\tilde{r},C_d) = \text{interp1d}\left( \tilde{r}, C_d \right)(\tilde{r'}) \).
% Each resampled joint position \( \mathbf{n'}_i \in \mathbb{R}^3 \) is then computed as   \({n'}_i = [ \hat{f}_x(\tilde{r'}_i), \hat{f}_y(\tilde{r'}_i), \hat{f}_z(\tilde{r'}_i)] \), for \(i = 1, ..., M\). 
% These resampled joint positions are then collected for each  kinematic chain:
% \( \mathbf{C}'(t) = [ \mathbf{n}'_1, \mathbf{n}'_2, \dots, \mathbf{n}'_{M} ] \) where \( M = 64 \).
%This ensures that the skeletal motion maintains the original shape while enhancing the number of spatial features per time step t. 
We concatenate the 5 interpolated chains $\mathbf{C}'(t)$ to create a skeletal motion image, defined as \(\mathbf{I} \in \mathbb{R}^{(M \times 5) \times T \times 3}\). 
% Concatenating the kinematic chains together allows us to represent the tensor in a format similar to RGB images, which are represented as \(Height \times Width \times [R,G,B]\), as seen in ~\ref{fig:motion_patches}c. This is convenient, since vision transformers are commonly pre-trained on RGB images. 

 % For multi-class prediction, words and integers are considered prediction classes. We use the integers ranging from 1 to 30, and integers are assigned a weight = 1.0. The total number of classes is then \( |\mathcal{V}| + 30\).
\subsection{Multi Class Labeling} \label{multiclass}

Labels for exercise types can be quite varied.  While some exercises are descriptively labeled, for example ``reverse lunge with right leg forward'', other exercises have acquired colloquial descriptions or standalone names such as ``Arnold press'' or ``birddog''. 
A core action (e.g., lunge) can be modified with an adverb (slow lunge), an attributive noun (skater lunge), a spatially related descriptor (right leg lunge) or an adjective (reverse lunge). In addition, there are compound exercises with conjunctions (for example, ``clean and press'') and equipment-specific exercises (wall sit). 

We consider individual words as well as integers (from 1 to 30) as prediction classes. This allows for exercise detection, where one exercise belongs to at least one class, as well as repetition counting. Together, these prediction classes create a set of vocabulary words \( \mathcal{V} \), however, not all words are of equal importance. We categorize words as belonging to one of 9 categories, as shown in Table \ref{classes}, where each category is assigned an importance weight.

 \begin{table}

\centering
\caption{Word classes and associated weights.}
  \label{classes}
\begin{tabular}{|p{140pt}|p{40pt}|p{180pt}|}
% \begin{tabular}{|m{0.25\linewidth} | m{0.14\linewidth} | m{0.14\linewidth} |}
\hline
\textbf{Category Name} & \textbf{Weight} & \textbf{Examples} \\
\hline
Core Action  & $1.0$ & curl, squat, press, hinge \\ 
Spatial Adjective Variation  & $1.0$ & alternating, wide, supine \\ 
Noun Variation  & $1.0$ & burpee, clamshell, skater, diamond \\ 
Body Part Noun  & $1.0$ & bicep, hamstring, shoulder \\ 
Temporal Adverb Variation  & $0.4$ & slow, hold, eccentric, power \\ 
Equipment Noun  & $0.1$ & dumbbell, chair, wall, heavy \\ 
Combination Conjunction  & $0.5$ & and, plus, with \\ 
Uncategorized  & $0.0$ & wacky, truffle, vogue \\ 
Integers  & $1.0$ & 1, 2, 3... 30 \\ 

\hline

% \multicolumn{3}{p{280pt}}{GMM results on Variations broken down by ablation study.} 
\end{tabular}
 
\end{table}
% \begin{table}[h]
%     \centering
%     \begin{tabular}{ p{400pt}}
%         \begin{itemize}
%             \item {[W=1.0] Core Action: e.g. curl, squat, press, hinge}
%              \item {[W=1.0] Spatial Adjective Variation: e.g. alternating, wide, supine}
%              \item {[W=1.0] Noun Variation: e.g. burpee, clamshell, skater, diamond}
%              \item {[W=1.0] Body Part Noun: e.g. bicep, hamstring, shoulder}
%             \item{ [W=0.4] Temporal Adverb Variation: e.g. slow, hold, eccentric, power }
%              \item{ [W=0.1] Equipment Noun: e.g. dumbbell, chair, wall, heavy}
%              \item {[W=0.5] Combination Conjunction: e.g. and, plus, with}
%              \item {[W=0.0] Uncategorized: e.g. wacky, truffle, vogue}
%              \item {[W=1.0] Integers: e.g. 1, 2, 3... 30}
%         \end{itemize}
%     \end{tabular}
%     \caption{}
%     \label{classes}
% \end{table}
% \( g_i \in \mathcal{G} \) where i = 1 to 9, 

Consider a dataset set which contains text labels. Each label \( L_i \) consists of a set of words: \(L_i = \{ w_1, w_2, \dots, w_m \}\), where \( m \) is the number of distinct words in label \( L_i \). Define the vocabulary \( \mathcal{V} \) as the set of all unique words in the dataset: \( \mathcal{V} = \bigcup_{i} L_i\). Each category is assigned a weight; we use the set of weights  \( \mathcal{W} = \{0, 0.1,0.4, 0.5, 1\} \). Each word \( w \in \mathcal{V} \) is assigned a weight, \(f: w \to \mathcal{W}\),  according to the category to which it belongs. All categories and their corresponding words can be found in the Supplemental Information Section \ref{supplemental}.

% via mapping \( c: \mathcal{V} \to \mathcal{G} \). Each category \( g \in \mathcal{G} \) is assigned a weight \( h: G \to \mathcal{W} \). We define the composite weight function \( f = h \circ c: \mathcal{V} \to \mathcal{W} \), so that  \( f(w) = h(c(w)) \). 

% The naming convention of physical exercises is subjective, where multiple names can be given to the same exercise. For example, an "overhead press" is equivalent to a "shoulder press". Barbell row vs. barbell dead row, prisoner squat vs. squat. We observe that the natural language labels given to human exercises can be categorized into the following major classes.

 % All major classes are assigned a weight of 1.0. 

% The four major categories have a weight of \( 1.0 \) and the four minor categories have a weight \( <1.0 \):

% \[
% w(K_i) =
% \begin{cases}
% 1.0, & \text{if } major class or integer
% <1.0, & \text{otherwise}
% \end{cases}
% \]

% Each exercise name label is then multi-hot encoded, where a 470 bit vector is assigned a value of 1 at the corresponding label class indices. Our ground truth labels for exercise identification contain between 1 and 7 classes, with a median of 3. 
% : \( \mathcal{K} = \{ K_1, K_2, \dots, K_{|\mathcal{V}|} \} \)
 
Each word \( w \in \mathcal{V} \) is considered a unique prediction class. For multi-class prediction, we define the multi-hot encoded vector \( \mathbf{y}_i \) of size \( |\mathcal{V}| \) for label \( L_i \): \(\mathbf{y}_i \in \{0, 0.1,0.4, 0.5, 1\}^{|\mathcal{V}|} \), where:

\[
\mathbf{y}_{i, j} =
\begin{cases}
f(w_j), & \text{if } w_j \in L_i  \\
0, & \text{otherwise}
\end{cases}
\]
% \label{ce_loss}
% \text{ and } f(w_j) \in \{K_1, K_2, K_3, K_4\} \\

The model outputs a predicted vector \( \hat{\mathbf{y}}_i \), and the multi-class loss function is computed as:

\begin{equation}
\label{eq:ce_loss}
\mathcal{L}_i = -\sum_{j=1}^{|\mathcal{V}|} y_{i, j} \log \hat{y}_{i, j}
\end{equation}

This represents the categorical cross-entropy loss between the predicted and ground-truth multi-hot encoded vectors.

\subsection{Training and Model Fitting} \label{training}

 % It has shown by \cite{kim_vilt_2021} that encoders of visual images do not need to include complex CNN architectures, and instead, linear projection is sufficient. This architecture has lead to successfully aligning visual features to natural language descriptors. This architecture was pre-trained using masked language modeling approaches found in \cite{devlin_bert_2019, li_visualbert_2019}. 

\begin{figure}
    \includegraphics[width=\linewidth]{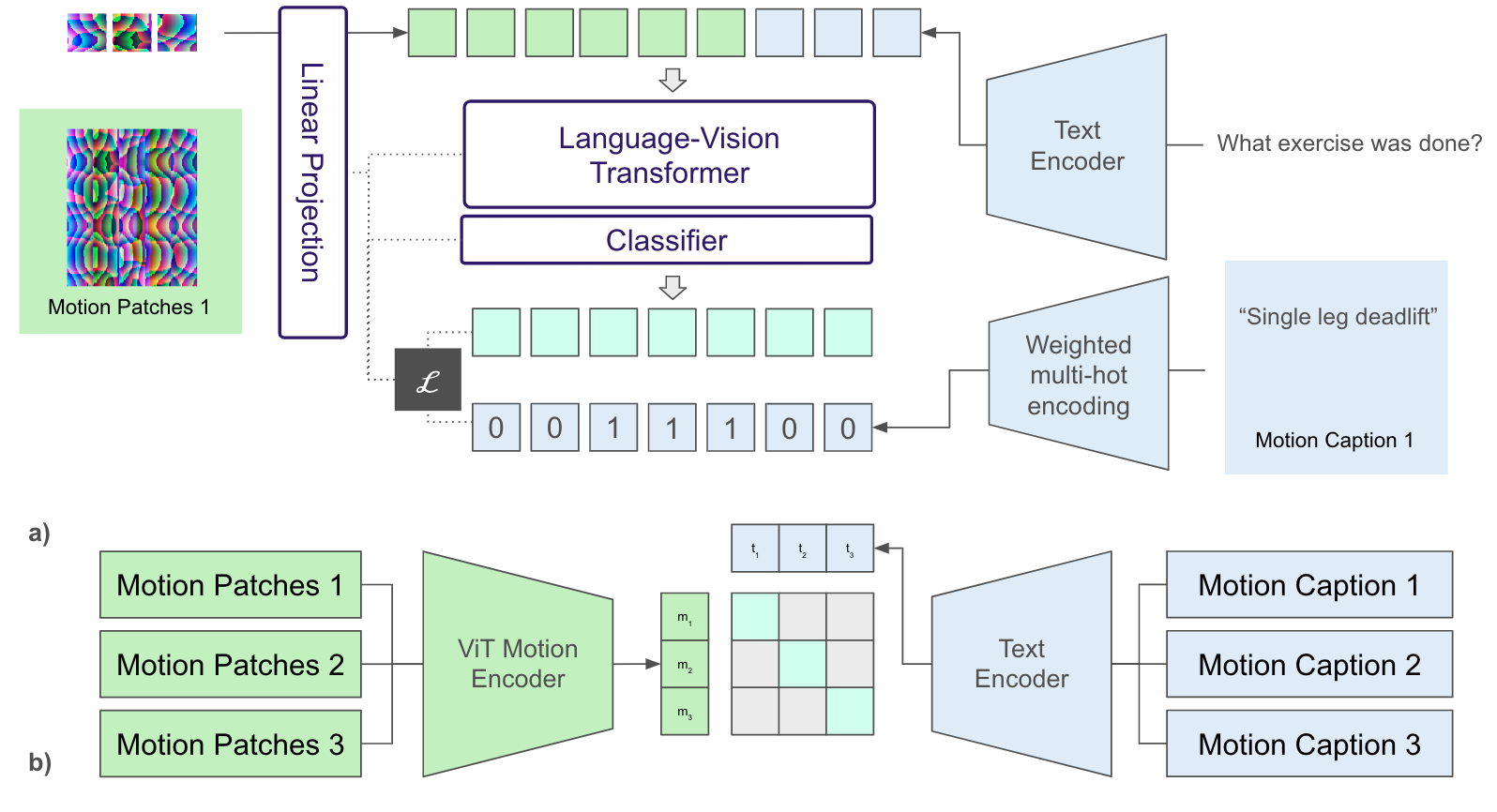}
    \caption{Comparing architecture types for motion-language understanding. a) shows our architecture which uses multi-class cross entropy for learning relationships between individual words and motion patches. b) shows the contrastive loss approach used by other works to learn global relationships between motions and language.}\label{fig:comparison} 
 %   \vspace{-5mm}
  \end{figure}

We use transfer learning, using an existing architecture ViLT \cite{kim_vilt_2021} with pre-trained weights, for learning the relationships between motion and language. ViLT was pre-trained on images along with their textual descriptions. We chose ViLT over the CLIP-style contrastive alignment used in \cite{yu_exploring_2024} because CLIP-based models primarily learn global relationships between images/motions and text. This can be seen in Fig.~\ref{fig:comparison}b.  Since CLIP pools cross-entropy losses at the image-text level, it does not explicitly model fine-grained correspondences between individual motion patches and specific words. In contrast, our modified ViLT directly processes motion patches and text within a shared transformer, allowing it to learn detailed relationships between sub-motions (kinematic chains) and linguistic components, as shown in Fig.~\ref{fig:comparison}a. This makes ViLT better suited for multi-task motion analysis, where localized motion-language alignment is critical.

In the ViLT architecture, text and image embeddings are concatenated before computing several layers of self-attention, which learn the interactions between the two modalities. To adhere to the ViLT maximum image size of $640 \times 384$, we temporally under-sample the motion patches at half their original frame rate (from 25 FPS to 12.5 FPS). This allows for motions to be up to 51.2 seconds long. Motion images \( \mathbf{I} \) smaller than $640 \times 384$ are zero padded; \( \mathbf{I}_p \) represents the padded image.
%\(\mathbf{I}_p = \text{pad}(\mathbf{I}, 384, 640)\). 
% A full attention mask \( \mathbf{M} \) of the same size is applied: \(\mathbf{M} \in \{0,1\}^{384 \times 640}
% \). 

We train the multi-class classifier head from scratch. We initialize the 471 classes with the vocabulary \( \mathcal{V} \), made up of 441 unique words representing 1,749 exercise labels and the integers between 1 and 30 representing possible repetition counts. During training, the inputs to the model are a padded motion image \(\mathbf{I}_p \) with corresponding task-related question as defined in Section \ref{olympia}. 
% The text is encoded using WordPiece \cite{song_fast_2021}. The motion images are sliced into patches of size [32,32,3], where each patch encodes a 2.56-second sub-motion segment. This is sufficient for capturing most human exercise cycles, which typically range from 1.5 to 4 seconds. Each motion patch encoded using a trainable linear projection layer \cite{kim_vilt_2021}
The learnable output is a multi-hot encoded vector as defined in Section \ref{multiclass}.
 
% These encodings are then added to positional encodings, modality encodings, and learning  concatenated

% Since an individual {\bf{subject}} may perform multiple sets of repetitions, our partitioning is done at the {\bf{set level}} to ensure diverse representation. Each exercise in the dataset is represented in at least one training, validation, or test set. Because some exercises are only seen once in the entire dataset, the random splitting  {\bf{does not guarantee full coverage of all exercises in each split.}}

\begin{figure}
    \includegraphics[width=\linewidth]{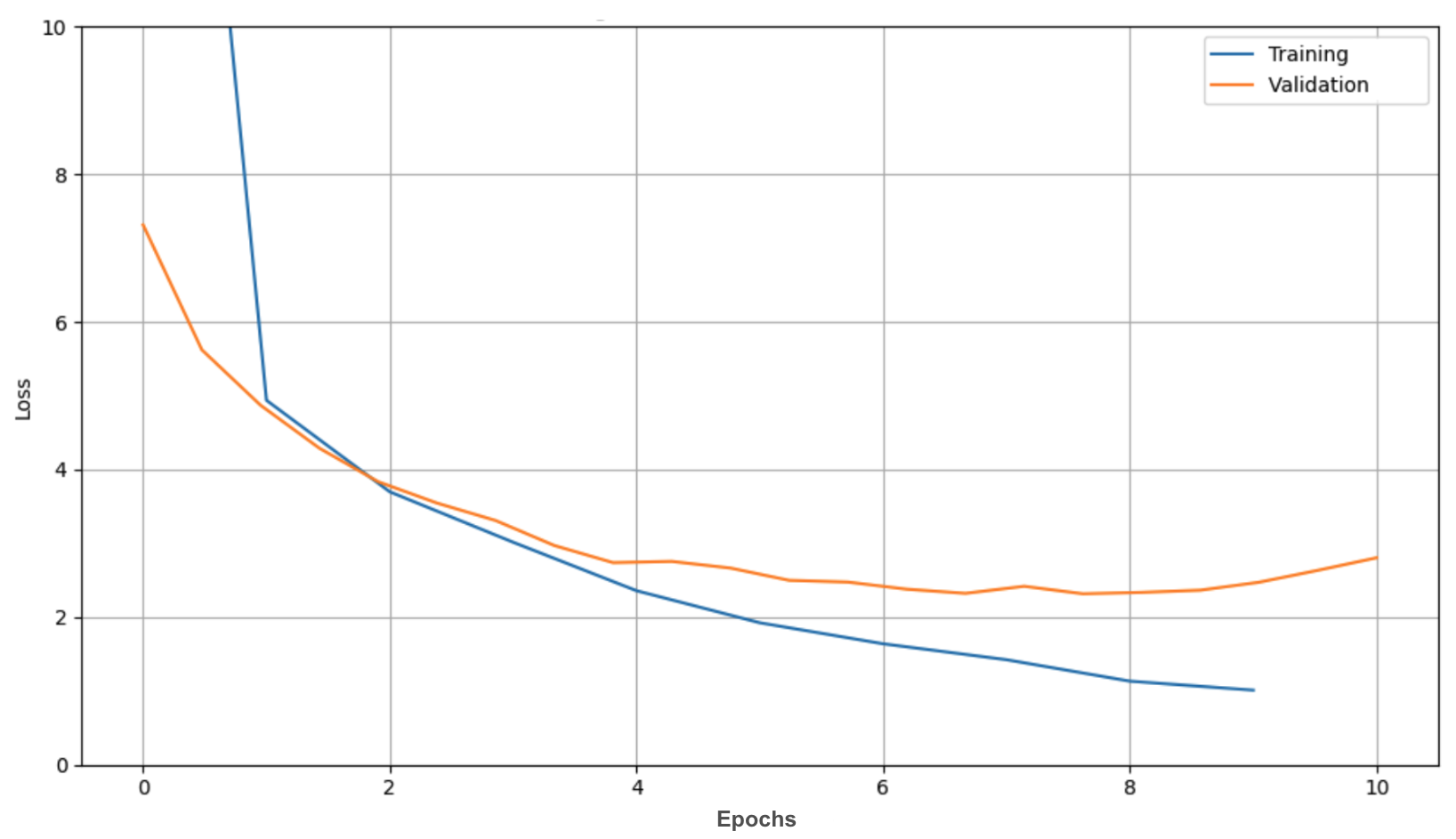}
    \caption{Training and Validation losses on the Olympia dataset.}\label{fig:loss} 
 %   \vspace{-5mm}
  \end{figure}

The dataset is randomly partitioned into $80\%$ training, $10\%$ validation, and $10\%$ testing videos. A full attention mask is used for the transformer model. Adam optimizer is used with early stopping (patience of 5 epochs). The learning rate is set to $5 \times 10^{-5}$ with a batch size of 32.  The resulting loss function for the multi-class cross-entropy loss is seen in Fig.~\ref{fig:loss}, showing the model learns well on the dataset. 

\subsection{Dataset}
\label{olympia}
Olympia consists of several existing datasets \cite{fieraru_aifit_2021, noauthor_papers_nodate-1, noauthor_mm-fit_nodate, aguilar-ortega_uco_2023, hu_transrac_2022} as well as open-source fitness videos. The open-source videos were filmed from various distances and angles, include various backgrounds and lighting scenarios. Combined, these video sources yielded 7,618 videos. 

{\bf {Existing Datasets:}}
Fit3D \cite{fieraru_aifit_2021} is a collection of 47 exercises recorded from 4 different camera angles. We  use the 8 subjects which have corresponding exercise label and repetition count data. Out of the 47 exercises, 17 were labeled as $warmup X$ exercises whose descriptions were not given. We keep the idiosyncratic names of $warmup X$ for the model, to test the ability of the model to learn global motion properties in addition to sub-motion properties. The remaining 30 exercises in Fit3D were repeated 5 times by each subject. To add repetition variability, we previously introduced a Variations dataset \cite{postlmayr_personalpt_2024}, where the repetition count for 6 different exercises was varied between 5-11 repetitions across subjects. These exercises included: knee diamond push up, diamond push up, tandem squats, half squat, chair squat, squat. InfiniteRep \cite{noauthor_papers_nodate-1} is a synthetic dataset which includes 100 videos for 10 exercises (birddog, arm raise, bicycle crunch, curl, fly, leg raise, overhead press, pushup, squat, superman). Our dataset from \cite{garg_short_2024} contains 8 subjects filmed from 5 camera angles performing 4 exercises (single leg romanian deadlift, romanian deadlift, single leg mini squat, external rotator cuff rotations. MM-Fit \cite{noauthor_mm-fit_nodate} containes 10 exercises: (squat, push up, dumbbell shoulder press, lunges, standing dumbbell rows, sit ups, dumbbell tricep extensions, bicep curls, sitting dumbbell lateral raises, jumping jacks). UCO \cite{aguilar-ortega_uco_2023} contains 10 exercises: Bending the knee without support while sitting Seated Right Lower, Horizontal weighted openings Standing Left Upper, Circular pendulum Standing Left Upper, Bending the knee without support while sitting Seated Left Lower, Bending the knee with support while sitting Seated Right Lower, Circular pendulum Standing Left Upper, Lift the extended leg Supine Right Lower, Bending the knee with bed support Supine Left Lower, Horizontal weighted openings Standing Right Upper, Shoulder flexion Seated Right Upper).

{\bf {Open-source:}}
We curated exercise videos from 7 individuals, ranging from 10-45 minutes each, totaling 2,273 videos. Our team annotated the exercise timestamps, names, and repetition counts. Each video contains only a single exercise set. Unlike previous datasets, the open-source fitness dataset included multiple combined movements into compound exercises, e.g. leg lunge + knee drive. 

We curated the natural language labels with the following steps. All verbs are converted to singular (e.g., squats to squat). Conjunctions are kept (e.g. +, \&, 'and'), but other than these conjunctions, special characters were removed (e.g., Child's pose) and hyphens are removed (e.g., push-up to pushup). All alphabetical labels are converted to lowercase and corrected for spelling. We do not convert to singular for anatomical classes (e.g., arms vs. arm) since they contain important information about the exercise. Abbreviations are expanded (L to left, rdl to romanian deadlift). Letters indicating posture are kept (w raise, t press, x plank), number abbreviations are corrected for (1-leg to single leg)

% We labeled the open-source videos by human annotation as well as using Google Gemini 2.0 to determine the start/stop of exercise sets in videos which could be up to 1:00hr long. We determine that Gemini would often hallucinate for the repetition counts, and we relied on human annotation for repetition count. 
% Ultimately, our dataset contains 9,000 videos with exercise names and repetition count. We follow the mirroring augmentation seen in HumanML3D \cite{guo_ericguo5513humanml3d_2025} resulting in a total of 18,000 videos.  In this mirror augmentation, the natural language labels of "left" and "right" are exchanged.

All Olympia videos have a corresponding motion caption label $L_i$, denoting the type of exercise, and a repetition count. For each video, we generate question-answer pairs corresponding to the two tasks i) exercise recognition and ii) repetition counting. The answer is the exercise name or repetition count of the video. We predefine several natural language variations of each task in the form of questions, e.g., ``Which exercise was performed?'' and ``How many repetitions were done?'' to create verbose question-answer pairing. These variations are shown in the Supplemental Information Section \ref{supplemental} Finally, we have 152,360 question-answer pairings for 7,618 unique videos. 

Each video is pre-processed, so that a video contains a set of repetitions for a single exercise. The video are converted into 3D keypoints by MotionBERT, which extracts 17 keypoints per video frame. The 17 keypoints follow the Human3.6M \cite{ionescu_human36m_2014} format. These human pose estimations are subsequently converted into motion images. 

% TODO: Potentially another experiment. How do you visualize that the model learned spatial/temporal info??
% In previous works, we have utilized the practice of goniometry (measuring joint angles), first and second time derivatives of angular data, as well as frequency estimation (FFT/DCT). We find this practice to be suitable for less complex machine learning models; however, due to the post-set estimation nature presented in our work, we believe that image encoder learns latent representations of these feature relationships. 
% We present more on this in our Experiments section
%  \ref{experiments}. 

\section{Results}

% wandt_repnet_2019
Our model performs two tasks: exercise identification and repetition counting. Here, we present the accuracy scoring methods used for each task.
For repetition counting, we follow the metric of off-by-one (OBO) and mean absolute error (MAE) found in other repetitive action counting work \cite{hu_transrac_2022, fieraru_aifit_2021}. Off-by-one is the fraction of test videos where the model prediction count $count_{pred}^{i}$ is within one repetition of the test video ground truth count $count_{gt}^{i}$:
\begin{equation}
\label{eq:obo}
OBO = \frac{1}{V} \sum_{i=1}^{V} [| count_{gt}^{i} - count_{pred}^{i} | \leq 1]
\end{equation}
where $V$ be the number of videos evaluated. 

Mean absolute error is the average absolute difference between the predicted counts and the ground truth. Since the model can predict multiple classes with varying confidence scores, we only consider the single most confident prediction class for repetition counting. 
\begin{equation}
\label{eq:mae}
MAE = \frac{1}{V} \sum_{i=1}^{V} | count_{gt}^{i} - count_{pred}^{i} |
\newline
\end{equation}

% Unlike \cite{hu_transrac_2022, fieraru_aifit_2021}, we do not divide the difference between prediction and ground truth in MAE by the number of ground truth repetitions as this introduces a weighted penalty dependent on the number of repetitions and complicates comparisons across datasets, due to their varying numbers of repetitions.

\begin{figure}
    \includegraphics[width=\linewidth]{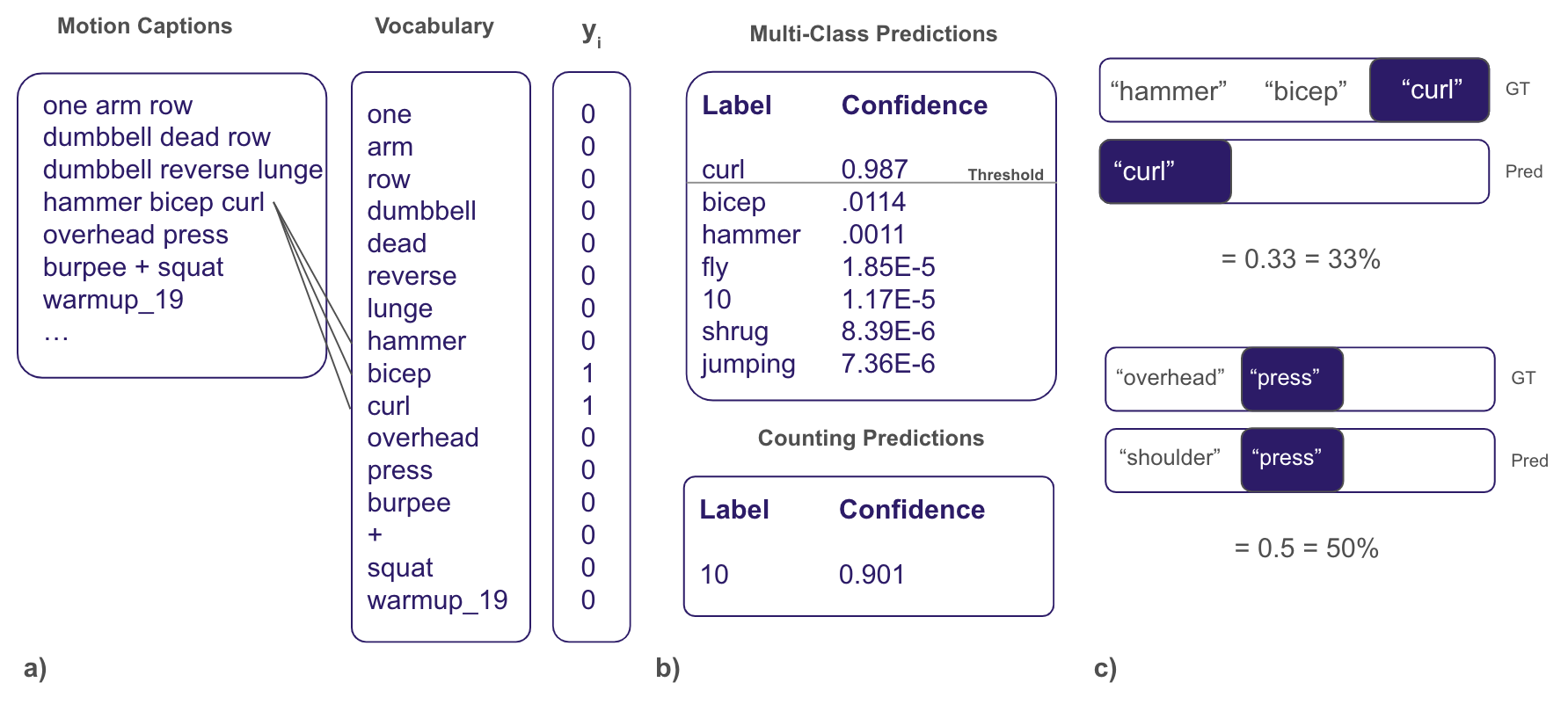}
    \caption{a) multi-class labeling using a shared Vocabulary and weighing words with major/minor category weights. For example, equipment words, such as "hammer", are weighted as 0 since the 3D HPE does not detect equipment. b) (top) shows confidence scores for multiple predicted classes which are considered for partial credit scoring (bottom) repetition counting only considers the most confident prediction. c) provides two examples of how partial credit accuracy is calculated for multi-class exercise detection.}\label{fig:n_gram} 
 %   \vspace{-5mm}
  \end{figure}

For exercise identification, we use the notion of ``partial credit" when scoring the accuracy of the model, using a metric similar to BLEU \cite{papineni_bleu_2002}, where we use class overlap. Given the model output logits \( \mathbf{z} \), we compute the softmax probabilities: \(\hat{\mathbf{y}}_i = \text{softmax}(\mathbf{z}_i)\), where each element \( \hat{\mathbf{y}}_{i,j} \) represents the prediction confidence for class \( j \) and associated word $w_j$. 
As shown in Figure \ref{fig:n_gram}b, only predictions \( \hat{\mathbf{y}}_j \) above the confidence threshold \(\tau = 0.05 \) are considered, such that:
% consider predicted classes which  extract the top-10 predictions sorted by confidence: \(\mathcal{P}_i = \{w_1, w_2, \dots, w_{10} \}, \)
%\quad \text{where } \hat{y}_{i,j_1} \geq \hat{y}_{i,j_2} \text{ for } j_1 < j_2\). Ultimately, we consider predictions 
% and keep those with a confidence score above \( \tau = 0.05 \): 
\(\mathcal{P}_i = \{ w_j  \mid \hat{\mathbf{y}}_{i,j} \geq \tau \}\).

If \( G_i \) represents the ground-truth set of words for sample \( i \),
%\( G_i = \{g_1, g_2, \dots, g_m\}\). Define 
the number of matching words is \( N_i = |\mathcal{P}_i \cap G_i| \).  The exercise detection accuracy is computed as: \( \text{Accuracy}_i = \frac{N_i}{|G_i|} \), as shown in Figure \ref{fig:n_gram}c. The final dataset-level accuracy is averaged over all samples. 
% \(
% \text{Accuracy} = \frac{1}{|D|} \sum_{i=1}^{|D|} \frac{N_i}{|G_i|}
% \)

\subsection{Results and Discussion}
\begin{figure}
    \includegraphics[width=\linewidth]{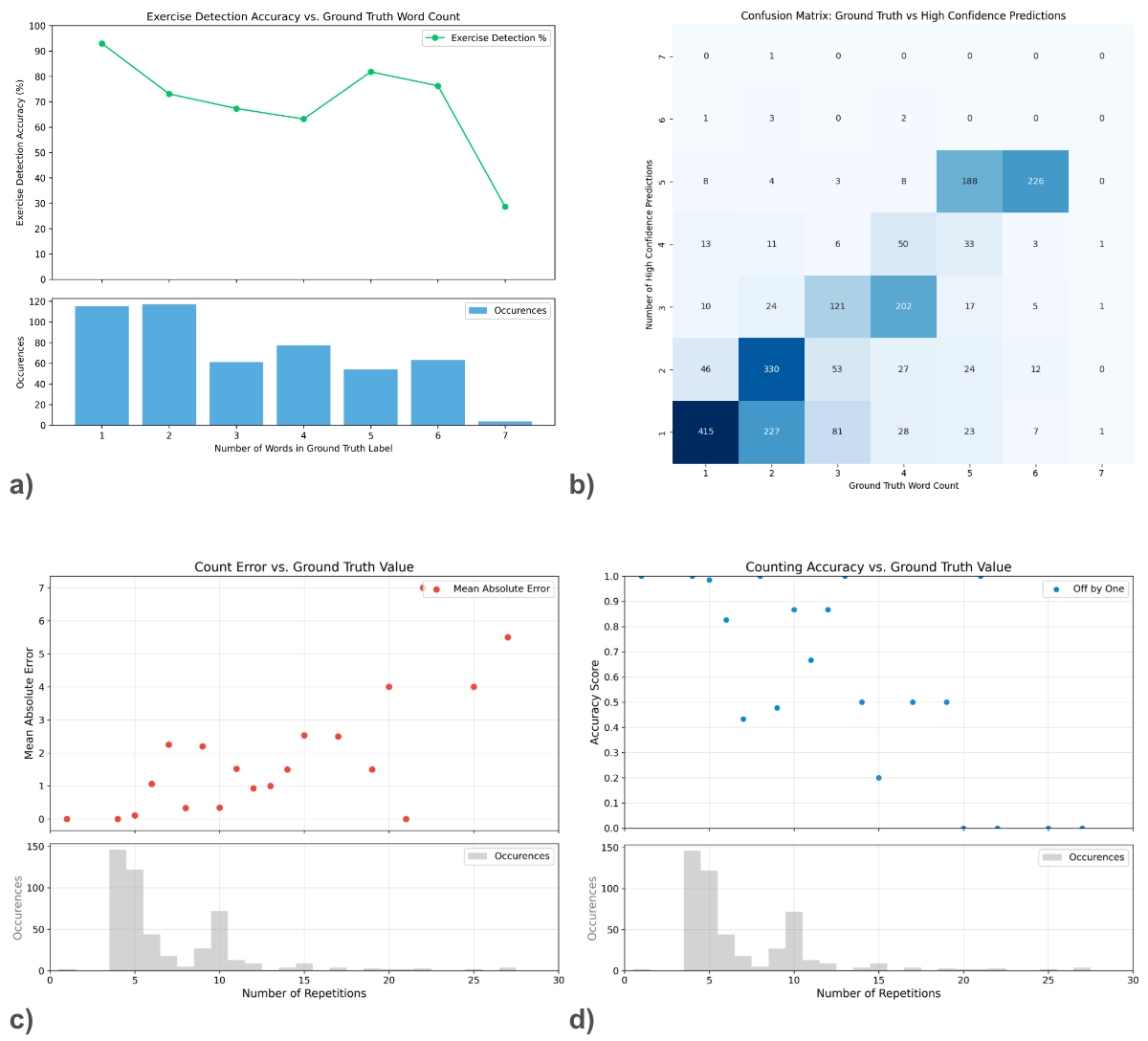}
    \caption{a) Comparing the accuracy of the model across the various exercise label word lengths in Olympia. b) The confusion matrix highlighting the number of words in the prediction label vs. the number of words in the ground truth label. c) Illustrating the model accuracy for repetition counting across various repetition counts. d) Illustrating the repetition counting absolute error for repetition counting across various repetition counts. }\label{fig:analysis}
 %   \vspace{-5mm}
  \end{figure}

We examine the accuracy of the model on exercise detection and repetition counting. Fig.~\ref{fig:analysis} delineates the distribution of data labels along with the corresponding model accuracies. To evaluate the model's performance across different motion caption lengths, we plot the exercise detection accuracy as a function of motion caption length in Fig.~\ref{fig:analysis}a. As anticipated, the model's accuracy declines with increasing caption lengths. The confusion matrix in Fig.~\ref{fig:analysis}b shows each the predicted and ground-truth numbers of words. Each column represents the number of predicted words on examples with only a fixed number of words, e.g. all examples containing only four ground truth words. The model tends to under-predict the number of words when compared to the ground truth lengths, even with a confidence threshold of $\tau = 0.05$, where the model could theoretically predict 20 words. Due to training the model equally on both tasks, exercise detection and repetition counting, at least half of the ground truth encoded vectors $\mathbf{y}_i$ only have a single non-zero element. Additionally, we illustrate the distribution of repetition counts within the dataset in Figure ~\ref{fig:analysis}c \& d. The significant concentration of videos with 5 and 10 repetitions is attributable to the common fitness practice of executing exercises in sets of 5 to 10 repetitions. 
We examine the accuracy of the model on exercise detection and repetition counting. Fig.~\ref{fig:analysis} delineates the distribution of data labels along with the corresponding model accuracies. To evaluate the model's performance across different motion caption lengths, we plot the exercise detection accuracy as a function of motion caption length in Fig.~\ref{fig:analysis}a. As anticipated, the model's accuracy declines with increasing caption lengths. The confusion matrix in Fig.~\ref{fig:analysis}b shows each the predicted and ground-truth numbers of words. If the ground truth label contains 4 words The model tends to under-predict the number of words when compared to the ground truth lengths, even with a confidence threshold of $\tau = 0.05$, where the model could theoretically predict 20 words. Due to training the model equally on both tasks, exercise detection and repetition counting, at least half of the ground truth encoded vectors $\mathbf{y}_i$ only have a single non-zero element. Additionally, we illustrate the distribution of repetition counts within the dataset in Figure ~\ref{fig:analysis}c \& d. The significant concentration of videos with 5 and 10 repetitions is attributable to the common fitness practice of executing exercises in sets of 5 to 10 repetitions. 
For counting absolute error, Figure ~\ref{fig:analysis}c shows an increasing absolute error on an increasing in repetition count. This can be expected. When more repetitions are performed in the same amount of time, as allowed by the maximum image size of the model, the signal can get deamplified.   Although the model scores s the number of repetitions increases, the OBO accuracy decreases.  accuracy gets less exact matches and more off-by-one scores. This is shown in the 5-10 rep region where the number of samples is high; however, the results get less reliable where repetitions $> 10$ due to less samples. 

For previously existing datasets included in Olympia, we break down the model exercise detection accuracy. The model achieves an exercise detection accuracy of: $98.77\%$ on UCO, $73.66\%$ on MM-Fit, $73.09\%$ on Fit3D, and $69.75\%$ on InfiniteRep. This shows that the model performs consistently across different exercise datasets. 

We qualitatively observe the following model performance. For peculiar exercise names, lower class overlap ratios was observed, e.g. "sumo squat pulse" or "burpee with 2 lunge jumps". Observed ambiguity in the video data may also lead to poor classification. For example, the model labeled a deadlift exercise as a "row" because the subject was seen picking up the barbell from the ground in a row-like fashion. Furthermore,there is room for improvement on the model, as the model was also observed to give entirely false classifications for some exercises. Here, the ground truth was distinctively different from the predicted class, e.g. leg lunge right curl vs. squat, left lunge arm press vs. squat. 
Surprisingly, the model was able to predict $warmup X$ exercises with an accuracy of $91.27\%$, validating that the model is able to learn global representations of motion in addition to using individual word classes to learn sub-motions.

\begin{table}
\centering
\caption[LiFT Results on Datasets]{
Results for LiFT Performance on  different datasets.}
 \label{tab:res} 
\begin{tabular}{|p{180pt}|p{70pt}|p{50pt}|p{50pt}|}
% \begin{tabular}{|m{0.25\linewidth} | m{0.14\linewidth} | m{0.14\linewidth} |}
\hline
\textbf{Results} & \textbf{Detection} & \textbf{OBO} & \textbf{MAE} \\
\hline
\textbf{Olympia} & \textbf{} & \textbf{} & \textbf{} \\

UCO  & $0.9877$ & $1.0$ & $0.0$\\ 
Open Source  & $0.7982$ & $0.687$ & $1.50$\\ 
MM-Fit  & $0.7366$ & $0.979$ & $0.126$\\ 
Fit3D  & $0.7309$ & $0.997$ & $0.007$\\ 
InfiniteRep  & $0.6975$ & $0.779$ & $0.772$\\ 
RepCount  & $0.6078$ & $0.593$ & $2.167$\\ 
\textbf{Experiments} & \textbf{} & \textbf{} & \textbf{} \\
Fit3D (Fine-Tune)  & $0.89$ & $.996$ & $0.131$\\ 
RepCount (Few-Shot)  & $0.24$ & $.186$ & $4.526$\\ 
KIMORE (Zero-Shot) \footnotemark[2]  & $0.208$ & $-$ & $-$\\ 
UI-PRMD (Zero-Shot)  & $0.167$ & $.647$ & $1.892$\\ 
UI-PRMD (Few-Shot (40)) \footnotemark[3]  & $0.497$ & $-$ & $-$\\ 
UI-PRMD (Few-Shot (200)) \footnotemark[3]  & $0.745$ & $-$ & $-$\\ 
\hline

% \multicolumn{3}{p{280pt}}{GMM results on Variations broken down by ablation study.} 
\end{tabular}
\end{table}

\footnotetext[2]{KIMORE dataset has a single recorded motion for each exercise}
\footnotetext[3]{UI-PRMD has each exercise performed for 10 repetitions. During fine-tuning on this data, the model learns to always predict 10.}

\subsection{Fine-Tune, Few-Shot, and Zero-Shot}
To understand the ability of our model to generalize, we test the model on 4 datasets which are publicly available. Specifically, we want to showcase: fine-tuning on the model, where the model is trained specifically on a dataset with many examples before evaluation; few-shot learning, where only a limited amount of dataset examples are used to train the model before evaluation; and zero-shot learning, where the dataset is not included in the training data. We show fine-tuning on Fit3D because the dataset includes many examples of many exercises. We show few-shot learning on RepCount because the dataset has a limited amount of exercises (6 exercises) with many examples per exercise. We test zero-shot capability on UI-PRMD and KIMORE datasets because these datasets have limited exercises/examples and include physical rehabilitation exercises different from the fitness exercises seen in the Olympia training data.
 Fine-tuning our model only on Fit3D \cite{fieraru_aifit_2021}, we train the model on 1246 examples and test on 160 samples. The model achieves an exercise detection score of $89\%$ and repetition counting score of $99.6\%$ OBO with a MAE of $0.13$. We attribute the high repetition counting score due to all subjects performing 5 repetitions of all the exercises, so fine-tuning on this dataset lead to this singular class prediction. 

For few-shot learning on RepCount, the model is trained on 100 examples and tested on 800. The RepCount dataset consists of 7 different exercises: pushup, pullup, jumping jacks, squats, bench press, frontal (arm) raises, situps. We exclude pommelhorse because of a lack of consistent repetitions. The repetition counting accuracy is competitive with other reported scores \cite{hu_transrac_2022} at $18.9\%$ OBO with a MAE of $4.5$, and the exercise detection score was $24\%$. Qualitatively analyzing the RepCount dataset, we conclude that this lower performance is a data issue. The dataset videos often include edited videos with changing camera angles, showing multiple subjects performing the exercise, and many exercise modifications within a particular exercise class.

Testing the model on zero-shot abilities, we { \bf do not} alter the prediction class vocabulary \( \mathcal{V} \) to include new word labels. Without any prior training, the model would effectively ignore these new words if they were added to the prediction classes. We use the KIMORE, which contains 5 physical rehabilitation movements without repetition counts. The model achieves an exercise detection score of $20.8\%$. We also test on UI-PRMD, which contains 10 fitness exercises, and all exercises were done with 10 repetitions. The model achieves an exercise detection score of $16.7\%$ and $64.7\%$ OBO with $1.96$ MAE. To qualitatively assess the model, we report ground truth vs. prediction comparisons for UI-PRMD. The model generalizes well for some exercises: (1) deep squat vs. squat, (2) stand-to-sit vs. squat, (3) standing shoulder extension vs. arm press extension, (4) standing shoulder abduction vs. arm raise, (5) side lunge vs. side squat. Reasonably for: (6) standing shoulder scaption vs. arm raise fly, (7) standing shoulder internal external rotation vs. arm raise, (8) inline lunge vs. single squat. Poorly for some: (9) hurdle step vs. arm, (10) standing active leg raise vs. arm. Exercises 9 and 10 were done with raised arms. 

We further investigate the effects of fine-tuning on UI-PRMD by providing training examples to the model. When the model is given 40 training examples and evaluated on 320, the exercise detection increases to $49.5\%$. When the model is given 200 training examples and evaluated on 160, the exercise detection increases to $74.5\%$.

\subsection{Ablation and Substitution Studies} \label{ablation}
We use ablation or substitution in several parts of our algorithm to test the validity of each step. We show the performance comparison of the model in Table \ref{tab:res}.

\subsubsection{Single Class Prediction}
To show the capability of the model without utilizing multi-class prediction, we treat the entire exercise label as a single class. Each full label \( L_i \) is treated as a distinct class: \(\mathcal{C'} = \{ C_1', C_2', \dots, C_{|\mathcal{D}|}' \} \) where \( |\mathcal{D}| \) is the number of unique labels in the dataset. This results in a vocabulary \( \mathcal{V} \) consisting of 1,749 unique labels, which is then used to initialize the model classification head. For this experiment, we do not use "partial credit" with class overlap. The $61.8\%$ accuracy obtained, as shown in Table \ref{tab:res}, is based on all-or-nothing string matching. Hence, we call this score our "Baseline". As expected, the exercise detection accuracy is lower than the "partial credit" approach accuracy of $76.5\%$. Furthermore, the repetition counting accuracy is minimally impacted $82.9\%$ vs $85.3\%$. This is expected since the different classification approaches are intended to affect the exercise detection score; however, we speculate the slightly lower repetition counting accuracy may be due to the large vocabulary increase for potential prediction classes.

\subsubsection{3D HPE Comparison}

We test the ability of the model to perform on lower quality data, i.e. using Mediapipe for 3D human pose estimation from RGB video. We use Mediapipe version 0.10.20 \cite{bazarevsky_blazepose_2020} due to its potential to run on mobile applications. Here, we extract 3D HPE from the same videos used in Olympia. We utilize the 'full' model, which has been shown to process video at over 30 frames per second on mobile devices. Mediapipe does not detect the keypoints root, spine, and thorax, which are present in the Human3.6M format obtained from MotionBERT. Therefore, we calculate these keypoints by interpolating the midpoint between the right and left hip for the root joint, the  midpoint between the right and left shoulder for the thorax, and the  midpoint between the root and thorax shoulder for the spine key point. The subsequent processing steps are the same as used with MotionBERT keypoints described in Section ~\ref{motion_patches}. The resulting accuracies are shown in Table \ref{tab:res}.

Mediapipe is known to have a higher 3D HPE prediction error than MotionBERT. Based on this, we expect our LiFT model to have a lower prediction accuracy when trained on and evaluated with lower quality motion data obtained. When using Mediapipe motion, we observe a $14.5\%$ and $15.9\%$ decrease in accuracy for exercise detection and OBO repetition counting respectively. This aligns with our expectations; however, it reinforces the notion that a end-to-end mobile system is possible using our motion-language model. We leave this mobile implementation as future work and speculate that more accurate, mobile-friendly 3D HPE is needed.

% \subsubsection{Motion Patches}
% % B. ATTENTION VISUALIZATION
% II. NUM. INTERPOLATED FEATURES

\begin{table}
\centering
\caption[Substitution Results for LiFT]{
 Ablation and Substitution results for LiFT. }
 \label{tab:ablations} 
\begin{tabular}{|p{200pt}|p{50pt}|p{50pt}|}
\hline
\textbf{Exercise Identification} & \textbf{Accuracy} & \textbf{Deviation} \\
\hline
Baseline Single-Class  & $0.618$ & $-$\\ 
MotionBERT Multi-class  & $\mathbf{0.765}$ & $-$\\ 
Mediapipe Multi-class & $0.654$ & $-$\\ 
\hline
\textbf{Repetition Counting} & \textbf{OBO} & \textbf{MAE} \\
\hline
% PCA Feature Selection  & $0.52$ & $3.15$\\ 
Baseline Single-Class & $0.829$ & $1.11$\\
MotionBERT Multi-class  & $\mathbf{0.853}$ & $\mathbf{0.645}$\\ 
Mediapipe Multi-class & $0.717$ & $1.43$\\ 
\hline
\end{tabular}
\end{table}

\subsubsection{Limitations and Future Work}
While our approach demonstrates strong performance in multi-task motion analysis, several limitations remain. First, the model requires a complete set of repetitions to analyze an exercise, which may limit real-time feedback. Second, our 3D human pose estimation approach does not detect exercise equipment (e.g., dumbbells, resistance bands), which could be valuable for tracking adherence and monitoring long-term strength progression. Third, our model does not yet assess repetition quality, though we believe the current architecture could support this with additional training data. We speculate that higher-fidelity 3D pose estimations obtained from more accurate 3D HPE models could enable more granular analysis of partial movements.

Finally, the model needs to be further improved to be able to run on a mobile device, utilizing lower quality 3D HPE such as Mediapipe. The LiFT model is well suited for mobile devices, requiring 433 MB of RAM to store the model weights with single precision floating-point values. In contrast, the aforementioned large language model weights require 25.6 GB of RAM with single-point precision floating-point values. More studies are required to optimize efficient deployment on mobile devices. Future work will focus on real-time 3D HPE for mobile implementation, rep quality assessment, and improving generalization to unseen exercises. Addressing these limitations will further enhance the model’s applicability in AI-driven fitness tracking, making intelligent coaching more accessible, scalable, and effective.

\section{Conclusions}

In this work, we introduced a robust, multi-task motion analysis model capable of performing exercise detection and repetition counting at a scale far exceeding prior approaches. While existing methods are often constrained to a limited set of exercises, our model generalizes across hundreds of possible movements by leveraging multi-modal vision-language representations and large-scale online fitness data. Through rigorous evaluation on multiple publicly available datasets, we demonstrated strong performance in both seen and unseen exercise categories, highlighting the model’s adaptability and scalability.

Our results suggest that vision-language models can effectively bridge the gap in multi-task motion analysis, paving the way for real-world applications in at-home exercise tracking and intelligent fitness coaching. By requiring only RGB video, our approach enables a software-only, privacy-conscious solution that can be deployed on consumer smartphones, democratizing access to AI-powered fitness analytics.

 \bibliographystyle{ieeetr}
 % \bibliographystyle{spmpsci} 
 % \bibliography{personalpt}
 \bibliography{LiFT}

\section{Supplemental Information}
\label{supplemental}

% Major
    "Core Action": {"weight": 1.0, "words": ["squat", "lunge", "press", "push", "pull", "row", 
        "deadlift", "hinge", "raise", "jump", "hop", "fly", "thruster", 
        "step", "sit", "kneel", "kick", "swing", "scaptions", "curtsy", "plank", "crunch", "tap", 
        "jack", "rotation", "walking", "lean", "extension", "flexion", "curl", "kickback", "punch",
         "slam", "dip", "twist",  "march", "bridge", "clean",  "fold", "tabletop", "hug",  "run", "inhale", "exhale", 
         "abduction", "adduction", "roll", "uppercut",  "tuck", "flutter","breathe", "snatch", "skip", "squeeze",  "bend", "thrust", "lift", "rest",
         "shuffle", "shoot",  "jog", "hook", "crawl",  "flap", "elevate", "grip", "pump", "bike", "blast", "whack", "slide", "shrug",
         "pullover",  "getp", "rock", "rotate", "rise", "walk", "bow", "smash", "balance", "clap", "situp", "pullup", "pushup"]},
% Major
    "Spatial Adjective Variation": {"weight": 1.0, "words": ["left", "right", "narrow", "close", "wide", "staggered", "split", "in", "out","mini",
         "up", "down", "circle","circular", "pendulum",  "diagonal", "unilateral", "bilateral", "v", "x", "straight", "bent", "forward", "side", "other",  "back", "deep",
          "reverse", "backward", "frontal", "front",   "raise", "reach", "shift", "overhead", "over", "alternating", "upright", "standing","single-leg",
          "release", "pose", "t", "full", "half", "external", "internal", "inverted", "criss", "cross", "lateral", "point", "drive", "w", "neutral", "negative" ,
          "hollow",  "open", "little", "small", "big", "large", "downard", "through", "decline", "triangle", "under", "isolated", "squatting",
          "drop", "horizontal", "vertical", "seated", "facing", "l", "r", "low", "high",  "twisted", "iso", "inner", "outer", "distal", "proximal", "dorsal", 
          "hanging", "elevated", "single", "double", "top", "bottom" , "supine", "prone", "sliding", "behind", "lying", "incline", "parallel", "hyper",
          "fixed",  "flat", "seated", "across", "downward", "together", "extended", "bending", "one", "apart", "jumping", "twisting", "pulse"]},

% Major
    "Noun Variation": {"weight": 1.0, "words": ["goblet", "sumo", "deadbug", "bulgarian", "romanian", "hack", "bird", "superman", "goodmorning",
        "gorilla", "arnold",  "burpee", "climber", "pike", "child", "bear", "spider", "crab", "cat", "cow", "camel", "farmer", "diamond",
        "air", "bicycle", "rdl", "donkey", "quiet", "dolphin", "star", "clamshell", "fire hydrant", "saw", "mule", "scissor", 
        "hammer", "manmaker", "puppy", "rainbow", "skull crusher", "dog", "cobra", "seal", "commando", "needle", "thread",
        "skater", "clam", "wheel", "yoga", "warmup", "tandem", "swimmer", "mountain", "bulldog", "predator", "skier", "russian",  "can",  
        "surfer", "starfish", "jackknife", "halo", "frog","driving", "butterfly", "boxer", "runner", "cheat", "preacher",
        "concentration", "turkish", "sissy", "princess"   , "girl", "roman", "trifecta", "prisoner", "inchworm","boat", "pigeon", "man", "maker"
    ]}

% Major
    "Body Part Noun": {"weight": 1.0, "words": ["tricep", "arm", "leg", "core", "bicep", "shoulder", "knee", "toe", 
        "calf", "hamstring", "quad", "chest", "trap", "back", "deltoid", "hand", "heel",  "neck", "glute", "ab",
        "rotatorcuff", "butt", "ankle", "palm", "upper", "lower", "torso", "body", "tibialis", "groin", "hip", "oblique", "spinal",
        "booty", "serratus", "chin","lat", "foot", "wrist", "elbow", "thigh"]},

    "Warmups" :{"weight": 1.0, "words": ["warmup1", "warmup2","warmup3", "warmup4", "warmup5","warmup6","warmup7", "warmup8", warmup9","warmup10", "warmup11",
    
    "warmup12","warmup13", "warmup14","warmup15", "warmup16", "warmup17",
    
    "warmup18",  "warmup19"]}

% Minor
    "Temporal Adverb Variation": {"weight": 0.4, "words": ["slow", "fast",   "hold", "tempo", "eccentric", "isometric", "stretch", "seconds", "tempo", "pop", 
        "speedy", "static", "dynamic", "power", "iso", "part", "quick", ]},
            
% Minor
    "Equipment Noun": {"weight": 0.1, "words": ["smith", "cable", "sled", "trx", "resistance", "bodyweight", "barbell", "battlerope",
        "chair", "weight", "rope", "wall", "without", "with", "support", "supported", "bed", "couch", "bench","dumbbell", "band",
        "medicine", "kettlebell", "ball", "bosu", "box", "floor", "suitcase", "sky", "world", "machine", "bar", "heavy", "light", "board", "table" ]},
% Minor
    "Combination Conjunction": {"weight": 0.5, "words": ["and", "then", "+", "\&", "with", "combo", "with", "to", "the", "all",  "whole",  "while", "-"]},

% Minor
    "Uncategorized" :{"weight": 0.1, "words": ["vogue", "wacky", "variation", "truffle", "far", "away","way",  "around", "from", "off", ""]}

"PROMPTS": {
    "prompt": "Q: How many one arm rows were performed? A:",
    "prompt": "Q: Count the number of one arm rows in this video. A:",
    "prompt": "Q: What's the total count of one arm rows? A:",
    "prompt": "Q: How many repetitions of one arm rows did you observe? A:",
    "prompt": "Q: Number of one arm rows performed: A:",
    "prompt": "Q: Tell me the count of one arm rows. A:",
    "prompt": "Q: What exercise is being performed? A:",
    "prompt": "Q: Name this exercise: A:",
    "prompt": "Q: Which movement is shown in the video? A:",
    "prompt": "Q: Identify the exercise being demonstrated: A:",
    "prompt": "Q: What type of exercise is this? A:",
    "prompt": "Q: Tell me the exercise being performed: A:",
}

%% else use the following coding to input the bibitems directly in the
%% TeX file.

% \begin{thebibliography}{00}

% %% \bibitem{label}
% %% Text of bibliographic item

% \bibitem{}

% \end{thebibliography}
\end{document}